\documentclass[a4paper, 10pt, conference]{ieeeconf}      
\usepackage{FG2025}
\usepackage{url} 
\usepackage{graphicx}
\usepackage{amsmath}
\usepackage{amssymb}
\usepackage{booktabs}
\usepackage{times}
\usepackage{epsfig}
\usepackage{subcaption}
\usepackage{float}
\usepackage{graphicx}
\usepackage{amsmath}
\usepackage{amssymb}
\usepackage{caption}
\usepackage{bbm}
\usepackage{subcaption}
\usepackage{amsmath}
\usepackage[T1]{fontenc}
\usepackage{multirow} 
\usepackage{booktabs}
\usepackage{balance}

\usepackage{fancyhdr}
\FGfinalcopy 

\IEEEoverridecommandlockouts                              
\def\FGPaperID{****} 

\title{dc-GAN: Dual-Conditioned GAN for Face Demorphing \\ From a Single Morph}

\author{\parbox{16cm}{\centering
    {\large Nitish Shukla and Arun Ross}\\
    {\normalsize
   Michigan State University, East Lansing, USA }
    }}

\begin{document}

\ifFGfinal
\thispagestyle{empty}
\pagestyle{empty}
\else
\author{Anonymous FG2025 submission\\ Paper ID \FGPaperID \\}
\pagestyle{plain}
\fi
\maketitle

 \thispagestyle{fancy}

\fancyfoot[L]{979-8-3315-5341-8/25/\$31.00 \copyright 2025 IEEE}

\begin{abstract}

A facial morph is an image strategically created by combining two face images pertaining to two distinct identities. The goal is to create a face image that can be matched to two different identities by a face matcher. Face demorphing inverts this process and attempts to recover the original images constituting a facial morph. Existing demorphing techniques have two major limitations: (a) they assume that some identities are common in the train and test sets; and (b) they are prone to the morph replication problem, where the outputs are merely replicates of the input morph. In this paper, we overcome these issues by proposing dc-GAN (dual-conditioned GAN), a novel demorphing method conditioned on the morph image as well as the embedding extracted from the image. Our method overcomes the morph replication problem and produces high-fidelity reconstructions of the constituent images. Moreover, the proposed method is highly generalizable and applicable to both reference-based and reference-free demorphing methods. Experiments were conducted using the AMSL, FRLL-Morphs, and MorDiff datasets to demonstrate the efficacy of the method.

\end{abstract}

\section{Introduction}

Morphing, in the face recognition literature, refers to the blending of two or more face images to generate a composite image that matches the identities represented in the constituent images with respect to a face matcher. Such a composite image, known as a facial morph,  can be incorporated into a biometric-based identity document, thereby allowing multiple individuals to share the same document. This can be used to compromise the security of an application \cite{ref10,ref11}. Traditionally, morphs were created by extracting facial landmarks from face images and then combining the two face images \cite{ref23,ref21,ref50}. However, more recently, deep learning models have been remarkably successful in generating high-quality morphs, especially those based on generative paradigms like diffusion models \cite{ref9} and generative adversarial networks \cite{ref13,ref14,ref20}.

Morph Attack Detection (MAD) is essential for retaining the integrity, security, and reliability of a biometric system \cite{ref10}. Current MAD techniques can be broadly divided into two categories: i) reference-based differential-image techniques \cite{ref16,ref53}, and ii)  reference-free single-image techniques \cite{ref18,ref51,ref41}. Reference-based methods use two images: the input face image (say, on the passport) and a reference face image (say, the live image of a person acquired at the port of entry). Reference-free methods use only one image: the input face image.  Although MAD techniques can reliably flag facial morphs, they do not reveal the images used to create the morph. From a forensic standpoint, it is valuable to recover images of the constituent identities\footnote{The constituent images are sometimes referred to as ``bonafides" (BF) in the literature, since they are real images of real people} from a morph flagged by MAD software. 
This process of extracting constituent face images from a morph image is called demorphing. Limited work exists in this field, especially in the reference-free category. In this paper, we focus on both reference-based as well as reference-free face demorphing. 
\begin{figure}
    \centering
    \includegraphics[width=\columnwidth]{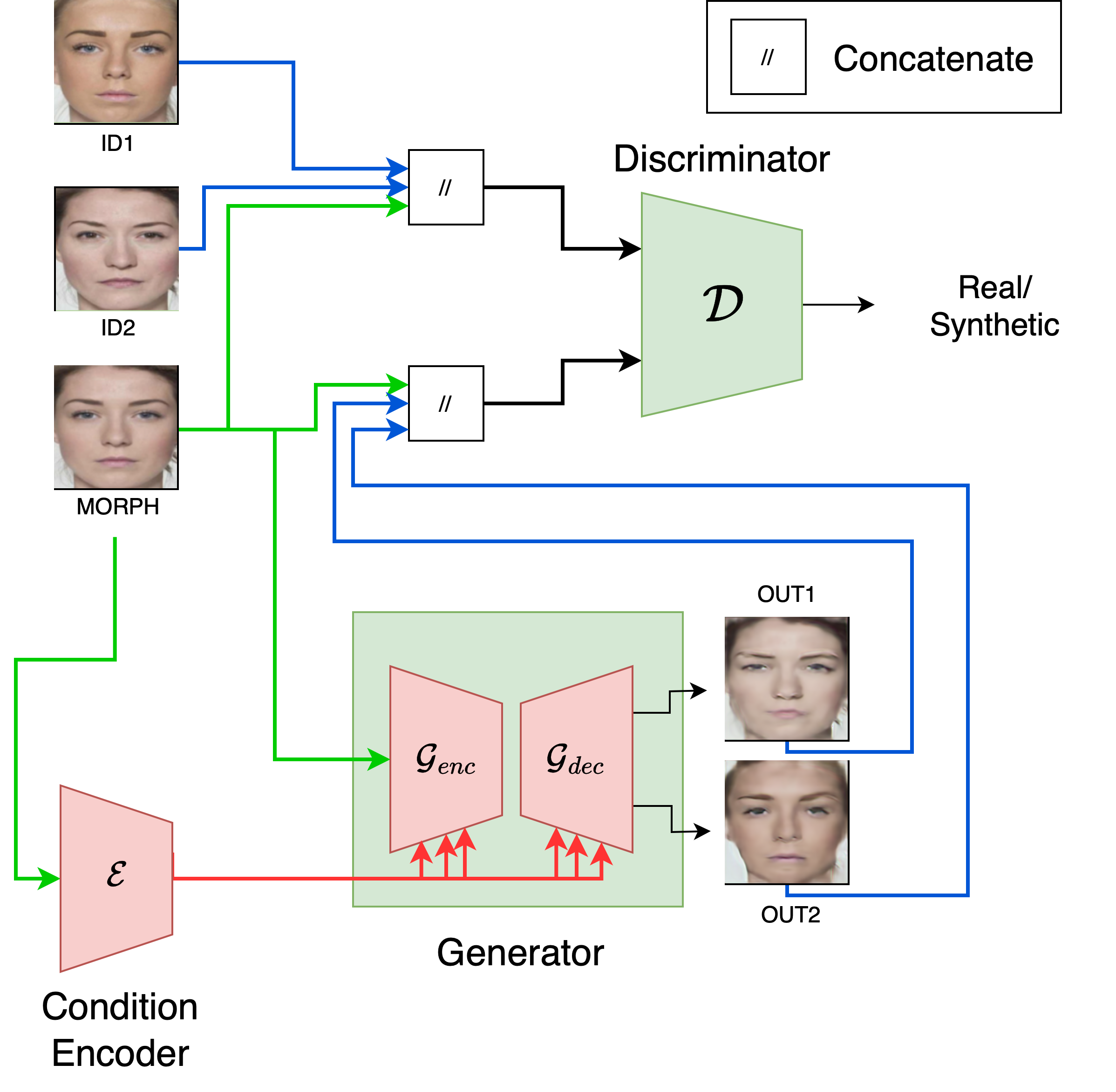}
    \caption{Dual-Conditioned GAN for Reference-Free Demorphing: An image encoder, $\mathcal{E}$, encodes the morph image, which is then used to condition the generator. The generator, based on a UNet architecture, $\mathcal{G}$, takes in the MORPH image and the encoded representation, $\mathcal{E}(MORPH)$, producing two outputs, OUT1 and OUT2. The discriminator is trained to distinguish between the real set (MORPH, BF1, BF2) and the synthetic set (MORPH, OUT1, OUT2), differentiating real from synthetic pairs.}
    \label{fig:arch}
\end{figure}
To set up the problem, we first identify multiple scenarios that are plausible in the real world. In the context of this work, the choice of train and test data will define these scenarios.  To formally define the various scenarios arising in the real world, let $\mathcal{X}_{train}$, $\mathcal{Y}_{train}$ denote the training data corresponding to the morphs and bonafides, respectively. Also, let $\mathcal{Y}_{test}$ be the set of test morphs. Assume that all morphs in $\mathcal{X}_{train}$ are necessarily generated only using the identities in $\mathcal{Y}_{train}$.  
\begin{equation}
    \forall x\in \mathcal{X}_{train}, \mathcal{M}(y_1,y_2)=x \Rightarrow \{y_1,y_2\}\in \mathcal{Y}_{train} 
\end{equation}

Given a test morph, $x$, created using the constituent images, $y_1$ and $y_2$, the scenarios are categorized as follows: 1) train and test morphs are generated from the same pool of identities, i.e., $y_1 \in \mathcal{Y}_{train} $ and $y_2 \in \mathcal{Y}_{train}$; 2) each test morph is created using one face image from the training set and one from an unseen identity, i.e., scenario 2 asserts that either $y_1 \in \mathcal{Y}_{train} $ or $y_2 \in \mathcal{Y}_{train}$ but not both; 3) train and test morphs are generated from a disjoint pool of face images (and disjoint identities), i.e., the identities involved in creating the training morphs do not participate in the creation of test morphs and vice-versa. In other words, $\mathcal{Y}_{test} \cap \mathcal{Y}_{train} =\phi$. Note that the datasets used in this paper consist of a single face image per identity, with only neutral face images from the FRLL dataset being used to generate the test morphs. Consequently, the terms ``images" and ``identities" are sometimes used interchangeably.

Most of the previous work assumed scenario 1 implicitly \cite{ref18,ref66}, but this is a strong assumption hampering the generalizability of the demorphing technique. Limited research has been conducted on demorphing in scenario 3, and most techniques seem to merely replicate the morph in both the outputs \cite{ref51}. 

In this work, we propose a demorphing technique free of any assumption regarding the identities of images (scenario 3) and can be used for both reference-based and reference-free demorphing. .











In summary, our contributions are as follows.

\begin{itemize}
    \item We propose dc-GAN, a novel GAN-based technique designed to extract constituent images from a single morph. Our approach is well-suited for both reference-free and differential (i.e., reference-based) demorphing tasks. 
    \item We conduct a thorough analysis of the \textit{implicit bias} present in morphs and extensively benchmark publicly available face recognition models on demorphing in the presence of high frequency visible artifacts which arise during the demorphing process.
    \item To the best of our knowledge, our method is the first to overcome the issue of \textit{morph-replication} when test morphs are created using previously unseen identities (scenario 3). 

\end{itemize}
    
\section{Background}
Face morphing refers to the process of combining two face images corresponding to two distinct identities to produce another face image (called a morph) which has a high biometric similarity with both constituent identities, as assessed by an automated face matcher.  Typically, a  morphing operator, $\mathcal{M}$, acts upon two input bonafides, $\mathcal{I}_1$ and $\mathcal{I}_2$, to produce the morph $\mathcal{X}$:
\begin{equation}
    \mathcal{X}=\mathcal{M}(\mathcal{I}_1,\mathcal{I}_2)
\end{equation}
The goal of $\mathcal{M}$ is to ensure that $\mathcal{B}(\mathcal{I}_1,\mathcal{X})>\tau$ and $\mathcal{B}(\mathcal{I}_2,\mathcal{X})>\tau$, where, $\mathcal{B}$ is a biometric face matcher that produces a similarity score and $\tau$ is the similarity threshold.

Face demorphing is the inverse of the morphing process. Given a morph $\mathcal{X}$, the goal is to recover the face images used to produce $\mathcal{X}$. Initial work on demorphing were primarily reference-based, i.e., they required an image of one of the identities to recover the other \cite{ref16,9484355}. The authors also assumed prior knowledge about the morphing method used during the morphing process. FD-GAN \cite{ref17} is a reference-based method that uses a symmetric GAN architecture and attempts to recover an image of the first identity from the morphed input using an image of the second identity. To validate the effectiveness of the generative model, it then attempts to recover the second identity using the output of the first identity. More recently, reference-free demorphing techniques have also been proposed with some degree of success. In \cite{ref51}, the authors decompose the morph image into two output images using a GAN that is composed of a generator, a decomposition critic, and two markovian discriminators. In \cite{ref18}, the author proposes a diffusion-based method for scenario 1 that iteratively adds noise to the morph image and recovers the constituent face images during the backward process. In \cite{ref66}, the authors decompose the morph into multiple privacy-preserving components using an encoder, and then recover the constituent face images by weighting and combining these components with a decoder.

\section{Proposed Method}
\label{proposed}


Face demorphing is the inverse of the morphing problem. The complexity lies not only
in the absence of knowledge of the morphing technique used (landmark-based or deep learning based) but in the lack of constraints on the output space.  Given a morph $\mathcal{X}$, the goal of the demorphing operator, denoted as $\mathcal{DM}$ is to recover the constituent face images,
\begin{equation}
    O_1,O_2=\mathcal{DM}(\mathcal{X}),
\end{equation}
satisfying the following conditions: 

\begin{equation}
\label{eq3}
||\mathcal{B}(O_1)-\mathcal{B}(O_2))||^2 <\theta
\end{equation}
and,
\begin{equation}
\label{eq4}
    \max_{j\in\{1,2\}} \max_{\substack{k \in \{1,2\} \\ k \neq j}} (||\mathcal{B}(O_j)-\mathcal{B}(I_k)||^2,||\mathcal{B}(O_j)-\mathcal{B}(I_j)||^2) >\epsilon
\end{equation}
where, $\theta$ and $\epsilon$ are similarity thresholds. Eqn. (\ref{eq3}) requires the reconstructed outputs to look different from each other in order to avoid morph replication, while Eqn. (\ref{eq4}) requires each output to align with its corresponding ground truth image. 

In \cite{ref51}, the authors attempt to recover identities from a single image using a GAN \cite{ref32} that includes a generator for demorphing and three markovian discriminators. However, the method has a limitation in that the outputs not only closely resemble the morph image but also appear similar to each other - the morph replication problem.  This issue arises because the morphing operator $\mathcal{M}$ is a highly nonlinear  function, making demorphing an ill-posed problem. 
Although their method employs three discriminators to optimize output similarity and visual realism, the generator only receives the morph image, which provides insufficient guidance for the demorphing process. The morph image alone may not be adequate to recover both constituent face images.

Inspired by this work, our method also utilizes GAN, but with a key difference: we introduce conditioning on both the generator and discriminator. This conditioning provides a more robust guidance for the generation of constituent images. We illustrate the architecture of our method in Figure \ref{fig:arch}. A UNet \cite{ref19} generator, $\mathcal{G}$, based on an encoder-decoder architecture, inputs the morph input and generates two outputs. Simultaneously, an image encoder, $\mathcal{E}$, encodes the morph image and feeds the resulting embedding into the generator. This morph embedding is injected into the intermediate latent layers of the generator. We refer to our method as a dual-conditioned GAN (dc-GAN) conditioned on (i) the morph image in the image domain and (ii) morph image embeddings in the latent domain. The discriminator receives a ``real'' triplet, consisting of the morph image along with the ground truth images (bonafides), and a ``fake'' triplet, composed of the morph image and the two generated outputs. The discriminator is then trained in an adversarial fashion alongside the generator to distinguish the real triplet from the synthetically generated one. Note that we also condition the discriminator with the morph image, providing it with the context needed to distinguish the morph from the constituent ground truth images. The conditions work in tandem to ensure that morph replication does not occur.

For differential, i.e., reference-based, demorphing, we modify our pipeline slightly. The image encoder now encodes both the morph image and the reference image, concatenating them to provide guidance to the generator. The generator receives both the morph and reference images as input (with 6 channels) and outputs the reconstructed image (with 3 channels corresponding to RGB). The discriminator architecture and function remain unchanged.

\subsection{Loss Functions }
The loss function for dc-GAN can be expressed as follows:

\begin{multline}
    \min_{\mathcal{G}}\max_{\mathcal{D}} \mathcal{L}_{\text{dcGAN}}(\mathcal{D},\mathcal{G}) = \mathbb{E}_{(x,\textbf{y})\sim p_{\text{data}}(\cdot)}[\log \mathcal{D}(\textbf{y}|x)] \\
    + \mathbb{E}_{z\sim p_z(z),    (x,\textbf{y})\sim p_{\text{data}}(\cdot)       }[\log(1-\mathcal{D}(\mathcal{G}(z|x,\mathcal{E}(x))|x)]
\end{multline}

The generator is conditioned on the morph, $x$, and its representation, $\mathcal{E}(x)$. We denote the bonafide image pairs as \textbf{y}. Similarly, the discriminator is conditioned on the morph as well. We introduce noise exclusively through the dropout mechanism, which is applied to several layers of our generator during both training and test phases, following the approach suggested in \cite{ref72}.

However, previous approaches observed that generating face images without explicit guidance results in blurry and less discriminatory features \cite{ref73}. On this front, we add another component to the loss function, along with the standard dc-GAN loss, which guides the model to generate feature-rich faces. Another issue is the lack of order in the generator's outputs (outputs are generated in no specific order). To address this, we align the generator’s outputs with the ground-truth face images using the widely employed cross-road loss \cite{ref51}.

\begin{equation}
    \mathcal{L}_{cr}= 
    \begin{cases} 
          \begin{aligned}
\min &[\mathcal{L}_1({\mathcal{I}}_1,\mathcal{O}_1)+ \mathcal{L}_1({\mathcal{I}_2},\mathcal{O}_2),\\
      &     \mathcal{L}_1({\mathcal{I}_1},\mathcal{O}_2)+ \mathcal{L}_1({\mathcal{I}_2},\mathcal{O}_1)
    ],
\end{aligned}&  \ \text{if reference free;} \\
&\\
      \mathcal{L}_1(\mathcal{I},\mathcal{O}), &  \ \text{if differential,} \\
       
   \end{cases}
    \label{cr_loss}
\end{equation}
where, $\mathcal{L}_1$ is the standard per-pixel loss,  $\{\mathcal{I}_{1},\mathcal{I}_2\}$ are the ground truth face images and  $\{\mathcal{O}_{1},\mathcal{O}_2\}$ are the generated outputs from our method.
In the case of reference-free demorphing from a single morph, the model generates two images, which are compared against the two ground-truth face images. The cross-road loss matches the correct output to the ground truth by computing the $\mathcal{L}_1$ loss between all possible pairs (only two). Taking the minimum ensures that the correct match is selected. For differential demorphing, the cross-road loss simplifies to the $\mathcal{L}_1$ loss between the single output $\mathcal{O}$, and the single ground truth, $\mathcal{I}$. The final loss is then given by:
\begin{equation}
    \mathcal{L}= \mathcal{L}_{dcGAN}(\mathcal{G},\mathcal{D})+\alpha \mathbb{E}_{x\sim p(\mathcal{X})}[\mathcal{L}_{cr}(\mathcal{O},\mathcal{G}(x,\mathcal{E}(x)))]
    \label{loss}
\end{equation}
where, $\mathcal{L}_1$ is the standard per-pixel loss, $x\sim p(\mathcal{X})$ is the input morph and $\mathcal{E}$ is the image encoder. Throughout our experiments, we set $\alpha$=1.

\begin{table}[h]
\centering
\caption{Dataset statistics after pre-processing. `L' refers to landmark based morphs, while `D' indicates deep-learning based morphs.}
\resizebox{\columnwidth}{!}{
\begin{tabular}{|l|r|r|r|r|r|}
\hline
\textbf{Dataset} & \textbf{Type}&\textbf{\parbox{2.0cm}{No. of train\\morphs}} & \textbf{\parbox{2.0cm}{No. of test\\morphs}} & \textbf{\parbox{2.0cm}{No. of train\\  subjects}} & \textbf{\parbox{2.0cm}{No. of test\\ subjects}} \\ \hline
AMSL    & L        & 823              & 302             & 56                 & 33               \\ \hline
StyleGAN   &D     & 480              & 161             & 60                 & 38               \\ \hline
WebMorph    &L    & 480              & 160             & 60                 & 37               \\ \hline
FaceMorpher  &L   & 249              & 100             & 55                 & 31               \\ \hline
MorDiff      &D   & 195              & 62              & 43                 & 26               \\ \hline

        OpenCV &L& 474 &160&60&37 \\ \hline
\end{tabular}
}
\label{tab:datasets}
\end{table}

\subsection{Implementation details}
\textbf{Demorphing:} Our pipeline consists of three components: i) an image encoder $\mathcal{E}$, ii) a generator $\mathcal{G}$, and iii) a discriminator $\mathcal{D}$. For $\mathcal{E}$,  we use a CLIP-based \cite{ref74} image encoder to effectively process and represent facial information. The choice of CLIP over face-specific models is due to its general-purpose utility. Since morphs often contain high-frequency artifacts and overlapping features, a general-purpose encoder like CLIP is more robust and offers better representation than models trained solely on real faces. We use the CLIP-ViT-B/32 model as a morph encoder, which is based on the transformer architecture. The 32 in ViT-B/32 refers to the size of the image patches. In this model, each image is divided into 32$\times$32 patches. These patches are then linearly embedded into a sequence of tokens that are processed by the transformer layers. We use a pretrained CLIP model from the HuggingFace library, which produces embeddings of size 1$\times$512. We repeat the embedding 77 times, mimicking the practice used in Stable Diffusion \cite{ref75}. For the generator, we use HuggingFace implementation of conditional UNet, namely, $\texttt{UNet2DConditionModel}$, featuring 6 ResNet downsampling blocks and an equal number of ResNet upsampling blocks. Self-attention is implemented in the fifth downsampling block and the second upsampling block, allowing the model to focus on important spatial features. Along with the image, the UNet also receives morph embeddings, which are injected into intermediate layers of both the downsampling and upsampling blocks. This integration allows the outputs to retain context about the original morph image throughout the generation process. This UNet implementation also requires a timestep, which we have consistently set to zero in all our experiments, effectively eliminating its impact. Finally, our Discriminator is based on a CNN architecture consisting of four blocks, each containing a convolutional layer, InstanceNorm \cite{ref76}, and LeakyReLU. The discriminator concatenates the real and synthetic triplets along the channel axis and outputs a real/synthetic score.

\textbf{Face Recognition Models and Parameters:}
We thoroughly evaluate our demorphing method using five publicly available Face Recognition Models: AdaFace \cite{ref22}, ArcFace \cite{ref77}, VGG-Face \cite{ref78}, FaceNet \cite{ref79}, and OpenFace \cite{ref80}. We evaluate the performance of each model in the presence of high-frequency artifacts introduced during the generation process. While we benchmark five different models, we use ArcFace as the primary model, consistent with its use in \cite{ref51}.  Each model is employed with its default settings as provided by the authors. We use cosine similarity to compute the biometric match score. Training is conducted using Adam optimization \cite{ref60} with multi-GPU support via \texttt{accelerate} \cite{ref81}. The training parameters are as follows: number of epochs: 300, learning rate:  $10^{-4}$, dropout rate: 0.1, $\beta_1$: 0.5 and $\beta_2$: 0.999. 

\begin{figure}[h]
    \centering
    \begin{subfigure}{0.6\columnwidth}
        \centering
        \includegraphics[width=0.8\linewidth]{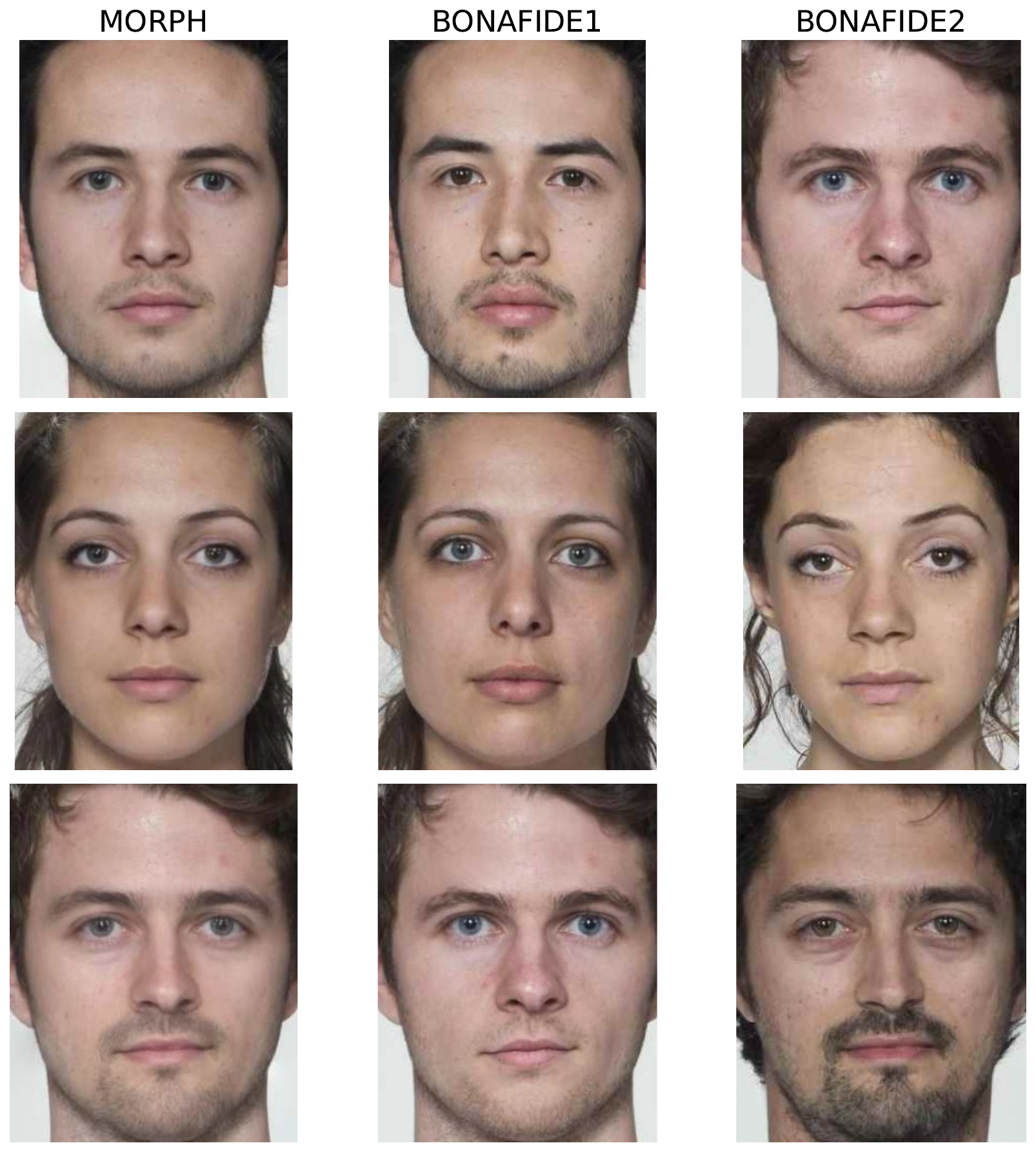}

    \end{subfigure}
    \vspace{0.5cm} 
    
    \begin{subfigure}{0.6\columnwidth}
        \centering
        \includegraphics[width=0.9\linewidth]{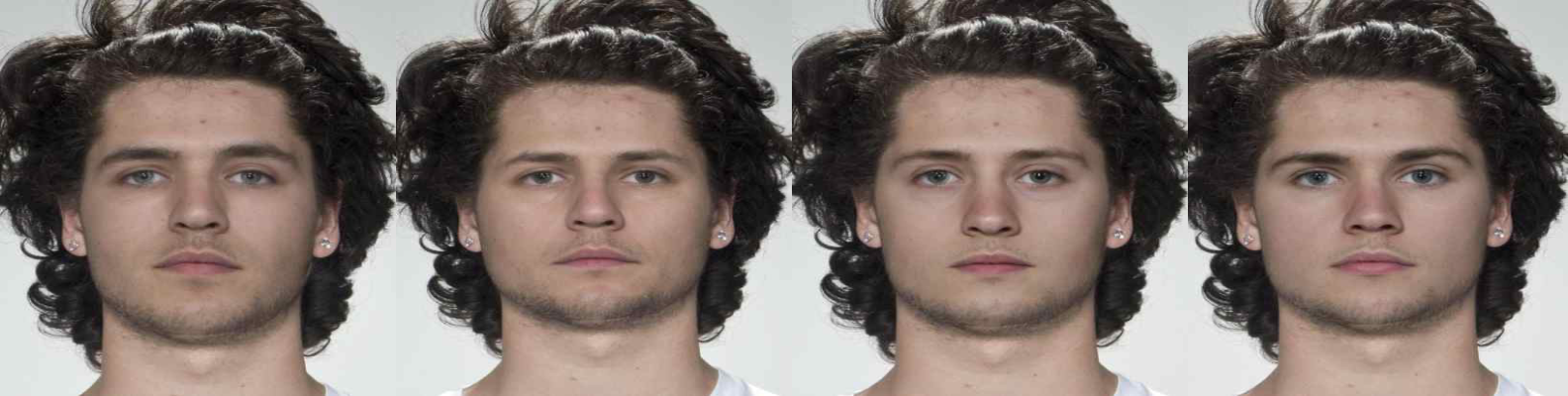}
        \label{fig:top-figure}
    \end{subfigure}
            \caption{(Top) Challenging samples in the AMSL-Morph dataset: Typically, morphs are created by overlaying a face image onto a \textit{base} image. These morphs are nearly indistinguishable from the base image, providing minimal information to the demorphing algorithm about the second image, thereby hindering the generation process. (Bottom) Morphs created from the same \textit{base image}.}
        \label{fig:hard-samples}
\end{figure}

\begin{table}[ht]
\centering
\caption{Intrinsic bias in morphs toward the \textit{base image}: We calculate the shift between MORPH-BF1 and MORPH-BF2 similarities and determine the $d'$ value. The morph images are biometrically closer to the \textit{base-image} than to the other image, making its recovery challenging.}
\resizebox{\columnwidth}{!}{
\begin{tabular}{|l|c|c|c|c|c|}
\hline
 & AdaFace & ArcFace & VGG-Face & Facenet & OpenFace \\
\hline
AMSL       & \textbf{2.26} & 1.39 & 1.42 & 1.15 & 1.71 \\
\hline
OpenCV     & 1.20 & 1.49 & \textbf{1.51} & 1.27 & 1.39 \\
\hline
FaceMorpher & 1.35 & \textbf{1.59} & 1.43 & 1.32 & 1.41 \\
\hline
WebMorph   & 1.05 & 1.29 & \textbf{1.41} & 1.15 & 1.48 \\
\hline
MorDiff    & 1.05 & 1.29 & 1.34 & 1.30 & \textbf{1.73} \\
\hline
StyleGAN   & 1.07 & 1.39 & 1.16 & 1.51 & \textbf{2.04} \\
\hline
\end{tabular}
}

\label{tab:dprime}
\end{table}

\section{Dataset and Preprocessing}
We conduct our experiments on three well-known morph datasets: AMSL \cite{ref64}, FRLL-Morphs \cite{ref65}, and MorDiff \cite{ref9}. The FRLL-Morphs dataset includes morphs generated using four different techniques: OpenCV \cite{ref67}, StyleGAN \cite{ref69}, WebMorph \cite{ref70}, and FaceMorph \cite{ref68}. In all three datasets, the bonafides are derived from the FRLL dataset, which comprises 102 identities, each with two frontal images (one smiling and one neutral). The morph counts in each of the datasets are as follows: AMSL: 2,175 morphs; FaceMorpher: 1,222 morphs; StyleGAN: 1,222 morphs; OpenCV: 1,221 morphs; WebMorph: 1,221 morphs; MorDiff: 1,000 morphs.

To split the dataset into training and testing sets, we follow an identity-disjoint protocol. We divide the 102 identities into training and testing sets using a 60-40 split. A morph image is included in the training (or test) set only if both bonafide identities used to create it belong to the training (or test) set. Morphs where one identity belongs to the training set and the other to the test set (or vice versa) are excluded. Additionally, any identities that do not participate in morph generation are also discarded. To maintain consistency across datasets, we only consider neutral faces (AMSL morphs are exclusively generated from neutral facial expression bonafides). 

\textbf{Intrinsic Bias in Morphs}: Figure \ref{fig:hard-samples} shows randomly selected examples from the AMSL dataset. Typically, a morph is created by superimposing a face image onto a \textit{base image}. This process inherently introduces a bias toward the \textit{base image}, making it extremely difficult to recover the other identity from the morph. To quantify this bias, we compute the similarity distributions between the morph and the two groundtruth images (MORPH-BF1 and MORPH-BF2) and calculate the $d'$ values for each in Table \ref{tab:dprime}. A higher $d'$ value indicates that the distributions are well separated. This method of creating morphs leaves very little visual information about the second identity, making it extremely challenging to reconstruct it accurately. As a result, the outputs often closely resemble the morph itself.

\textbf{Preprocessing}: All images are processed using MTCNN \cite{ref71} to detect faces, after which they are cropped to include only the face regions. The images are then normalized and resized to a resolution of $256\times256$. We discard the images in which faces cannot be detected. Importantly, no additional spatial transformations are applied to the images. This ensures that the facial features (such as lips, nose, etc.) of both morphs and ground-truth constituent images remain aligned during training. We present the dataset statistics in Table \ref{tab:datasets}.

\begin{figure}[h]
    \centering
    \includegraphics[width=1\linewidth]{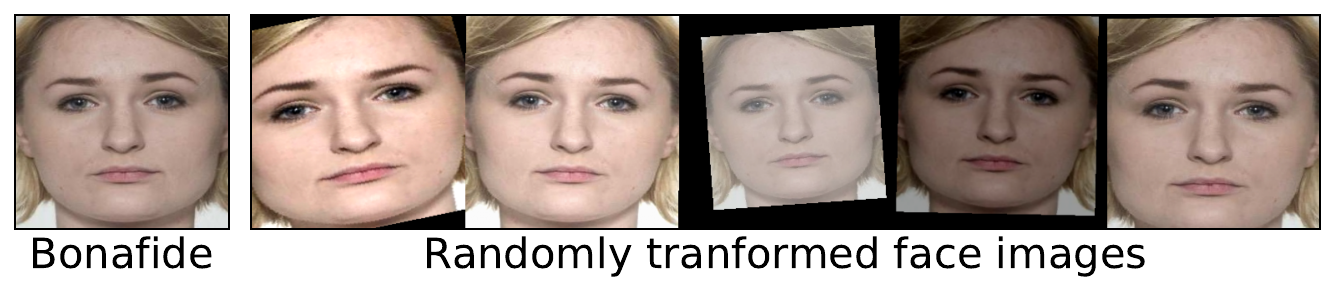}
    \caption{To simulate real-world variations, we apply random transformations to the reference image. It is important to note that these transformations are not applied during the training phase.}
    \label{fig:transformations}
\end{figure}

\section{Experimental Analysis}
\subsection{Evaluation Criteria}
Given a morphed face image MORPH as input, our reference-free method generates two output images, OUT1 and OUT2. Since the outputs are unordered, we determine the correct pairing with the ground-truth face images, BF1 and BF2,  by calculating the similarities for the two possible output-ground truth pairs. We use a face comparator $\mathcal{B}$ to assess facial similarity. If the sum $\mathcal{B}(\text{BF1}, \text{OUT1}) + \mathcal{B}(\text{BF2}, \text{OUT2})$ is greater than $\mathcal{B}(\text{BF1}, \text{OUT2}) + \mathcal{B}(\text{BF2}, \text{OUT1})$, we deem (BF1, OUT1) and (BF2, OUT2) as the correct pairs; otherwise, we swap OUT1 with OUT2. For each $\text{OUT}_i$, we consider its correct pair $\text{BF}_i$ as genuine, where $i,j \in$ \{1,2\} without replacement. The imposter score is computed by identifying the closest matching face image in the bonafide database, excluding the ground-truth image.

Note that this evaluation strategy enables us to detect \textit{morph-replication}. If our method replicates the MORPH as its outputs, i.e., $\mathcal{B}(\text{MORPH}, \text{OUT1}) \approx \mathcal{B}(\text{MORPH}, \text{OUT2})$, the similarity scores $\mathcal{B}(\text{BF1}, \text{OUT1}) + \mathcal{B}(\text{BF2}, \text{OUT2})$ will also be approximately equal to $\mathcal{B}(\text{BF1}, \text{OUT2}) + \mathcal{B}(\text{BF2}, \text{OUT1})$. As a result, the pairs (BF1, OUT1) and (BF2, OUT2) would be assigned arbitrarily, causing the match rate to drop close to zero.
We evaluate our reference-free demorphing method in terms of True Match Rate (TMR) at different thresholds of False Match Rate (FMR). 

For differential demorphing, we use a modified approach. Given a MORPH and a reference image REF, our method generates an output image, OUT. We treat the  ground-truth image BF and OUT as the genuine pair. For the imposter pair, we search the entire test set of bonafide images in the FRLL dataset and assign the face image with the third highest similarity to OUT, excluding GT and REF for obvious reasons (hence the third). During evaluation, we arbitrarily select one of the constituent face images used to create MORPH as the reference image. However, this does not reflect a real-world scenario, as the exact reference image used to create the morph might not be available during demorphing. To simulate this, we apply one or more random transformations to the reference image (see Figure \ref{fig:transformations}), such as Gaussian blur, jitter, affine transformation, etc. It is important to note that these transformations were not applied during training.

\begin{table*}[h]
\centering
\caption{Reference-Free Demorphing: We compute the True Match Rate (TMR in \%) at various False Match Rate (FMR in \%) thresholds (0.1/1/5/10) between the generated outputs (OUT1, OUT2) and the ground truth images (BF1, BF2) across the five FR models used. ArcFace performs most consistently across the various models used in the presence of high-frequency artifacts (See Figure \ref{fig:recon} and \ref{fig:diff-recon})  }.
\resizebox{\textwidth}{!}{%

\begin{tabular}{|c|c|c|c|c|c|}
\hline
\multirow{3}{*}{\textbf{Dataset}} & \multicolumn{5}{c|}{\textbf{TMR @ FMR $\uparrow$}} \\ \cline{2-6} 
                                  & \textbf{AdaFace} & \textbf{ArcFace} & \textbf{VGGFace} & \textbf{FaceNet} & \textbf{OpenFace} \\ \cline{2-6} 
                                  & 0.1/1.0/5.0/10.0 & 0.1/1.0/5.0/10.0 & 0.1/1.0/5.0/10.0 & 0.1/1.0/5.0/10.0 & 0.1/1.0/5.0/10.0 \\ \hline

AMSL& 17.75/42.91/46.93/56.83&
0.0/83.76/91.81/\textbf{93.86}&
0.0/18.25/28.50/43.34&
0.0/23.22/29.52/43.86&
1.02/5.34/9.04/13.65 \\
\hline
OpenCV & 5.7/15.82/41.77/51.58&
0.0/0.0/90.51/\textbf{93.99}&
0.0/0.0/31.01/50.63&
0.0/0.0/24.68/54.7/&
2.53/4.11/14.56/21.52 \\
\hline

StyleGAN& 1.27/1.59/6.69/8.6&
0.0/0.0/0.0/\textbf{58.92}&
0.0/0.0/0.0/7.3/&
0.0/0.0/0.0/6.69&
1.59/1.91/4.14/6.37 \\
\hline

WebMorph & 14.56/14.87/38.92/50.95&
0.0/0.0/0.0/\textbf{89.87}&
0.0/0.0/18.67/37.34&
0.0/0.0/10.13/33.86&
7.91/8.23/11.08/12.66 \\
\hline

FaceMorpher & 4.59/4.59/35.71/46.94&
0.0/0.0/29.45.0/\textbf{94.39}&
0.0/0.0/29.59/52.55&
0.0/0.0/9.69/58.67&
1.53/1.53/9.18/9.69 \\
\hline

MorDiff&21.43/21.43/35.71/47.32&
0.0/0.0/87.5/\textbf{93.75}& 
0.0/0.0/19.64/34.82&
0.0/0.0/22.32/27.68&
0.0/0.0/2.68/8.04 \\


\hline

\end{tabular}
}

\label{tab:performance_comparison}
\end{table*}

\begin{table*}[h]
\centering
\caption{Differential Demorphing: We compute the True Match Rate (TMR in \%) at various False Match Rate (FMR in \%) thresholds between the generated output (OUT) and the ground-truth image (GT) across the five FR models used.}
\resizebox{\textwidth}{!}{%

\begin{tabular}{|c|c|c|c|c|c|}
\hline
\multirow{3}{*}{\textbf{Dataset}} & \multicolumn{5}{c|}{\textbf{TMR @ FMR $\uparrow$}} \\ \cline{2-6} 
                                  & \textbf{AdaFace} & \textbf{ArcFace} & \textbf{VGGFace} & \textbf{FaceNet} & \textbf{OpenFace} \\ \cline{2-6} 
                                  & 0.1/1.0/5.0/10.0 & 0.1/1.0/5.0/10.0 & 0.1/1.0/5.0/10.0 & 0.1/1.0/5.0/10.0 & 0.1/1.0/5.0/10.0 \\ \hline

AMSL                          & 14.47/16.98/20.75/30.82&     86.88/87.50/88.75/\textbf{94.88}&     13.12/15.00/22.50/35.75&     7.50/11.88/24.38/32.50&     8.75/11.88/13.75/16.25 \\ \hline
OpenCV                          & 46.54/55.97/62.89/71.70&     73.12/76.88/83.12/\textbf{88.12}&     23.12/23.75/34.38/39.38&     32.50/32.50/46.25/57.50&     6.25/10.62/18.12/21.88 \\ \hline
StyleGAN                          &0.62/1.25/4.38/6.25&     36.02/41.61/45.34/\textbf{54.04}&     1.86/3.11/6.21/9.32&     3.11/3.73/5.59/8.07&     0.62/1.24/3.11/4.97 \\ \hline
WebMorph                          & 33.76/36.31/49.04/58.60&     67.50/81.25/87.50/\textbf{89.38}&     14.37/14.37/25.62/30.00&     26.88/27.50/38.12/44.38&     6.25/11.25/16.25/23.12\\ \hline
FaceMorph                          &79.00/79.00/94.00/\textbf{98.00}&     94.00/94.00/95.00/97.40&     34.00/34.00/53.00/58.00&     60.00/60.00/68.00/75.00&     9.00/9.00/23.00/26.00 \\ \hline
MorDiff                          & 73.77/73.77/77.05/81.97&     85.48/85.48/88.71/\textbf{90.32}&     40.32/40.32/40.32/43.55&     35.48/35.48/46.77/54.84&     17.74/17.74/20.97/27.42 \\ \hline
\end{tabular}
}

\label{tab:performance_comparison_diff}
\end{table*}

\begin{table*}[htb]
    \centering
    \caption{Restoration Accuracy: Comparison of our reference-free approach to SDeMorph \cite{ref18} and Identity-Preserving Demorphing (IPD) \cite{ref66}. Unlike our method, which operates on unseen faces (scenario 3) and is significantly more challenging, the other methods assume the same identities are present in both training and test morphs (scenario 1). In scenario 3, our method significantly outperforms the current state-of-the-art method. The scores for scenario 3 are provided by the original authors of \cite{ref18,ref66}.} 
    \resizebox{\linewidth}{!}{
    \begin{tabular}{|c|c|c|c|c|c|c|c|c|c|c|}
        \hline
        \multirow{3}{*}{Dataset} & \multicolumn{4}{c|}{Ours (scenario 3)} & \multicolumn{2}{c|}{SDeMorph (scenario 1)} & {SDeMorph (scenario 3)} & \multicolumn{2}{c|}{IPD (scenario 1)} & {IPD (scenario 3)} \\
        \cline{2-11}
        & \multicolumn{2}{c|}{AdaFace} & \multicolumn{2}{c|}{ArcFace} & \multicolumn{2}{c|}{AdaFace} & ArcFace & \multicolumn{2}{c|}{AdaFace} & AdaFace \\
        \cline{2-11}
        & Subject 1 & Subject 2 & Subject 1 & Subject 2 & Subject 1 & Subject 2 & Average & Subject 1 & Subject 2 & Average \\
        \hline

        AMSL & 8.60\% & 69.86\% & 85.43\% & 93.04\% & 97.70\% & 97.24\% & 12.56\% & 99.84\% & 99.56\% & 25.69\% \\
        FaceMorpher & 55.00\% & 43.00\% & 88.00\% & 93.00\% & 96.00\% & 99.50\% & 13.18\% & - & - & 37.82\%\\
        MorDiff & 70.96\% & 72.58\% & 83.87\% & 85.48\% & 78.00\% & 74.00\%     & 11.67 & - & - & 38.12\%\\
        StyleGAN & 0.00\% & 0.00\% & 34.16\% & 47.82\% & - & -                  & 0.00\% & - & - & 16.22\%\\
        WebMorph & 22.20\% & 33.75\% & 85.62\% & 90.62\% & - & -                & 12.80\% & - & - & 25.61\%\\
      
        \hline
    \end{tabular}
    }
    \label{tab:compare}
\end{table*}

\begin{figure}
    \centering
    \includegraphics[width=\linewidth]{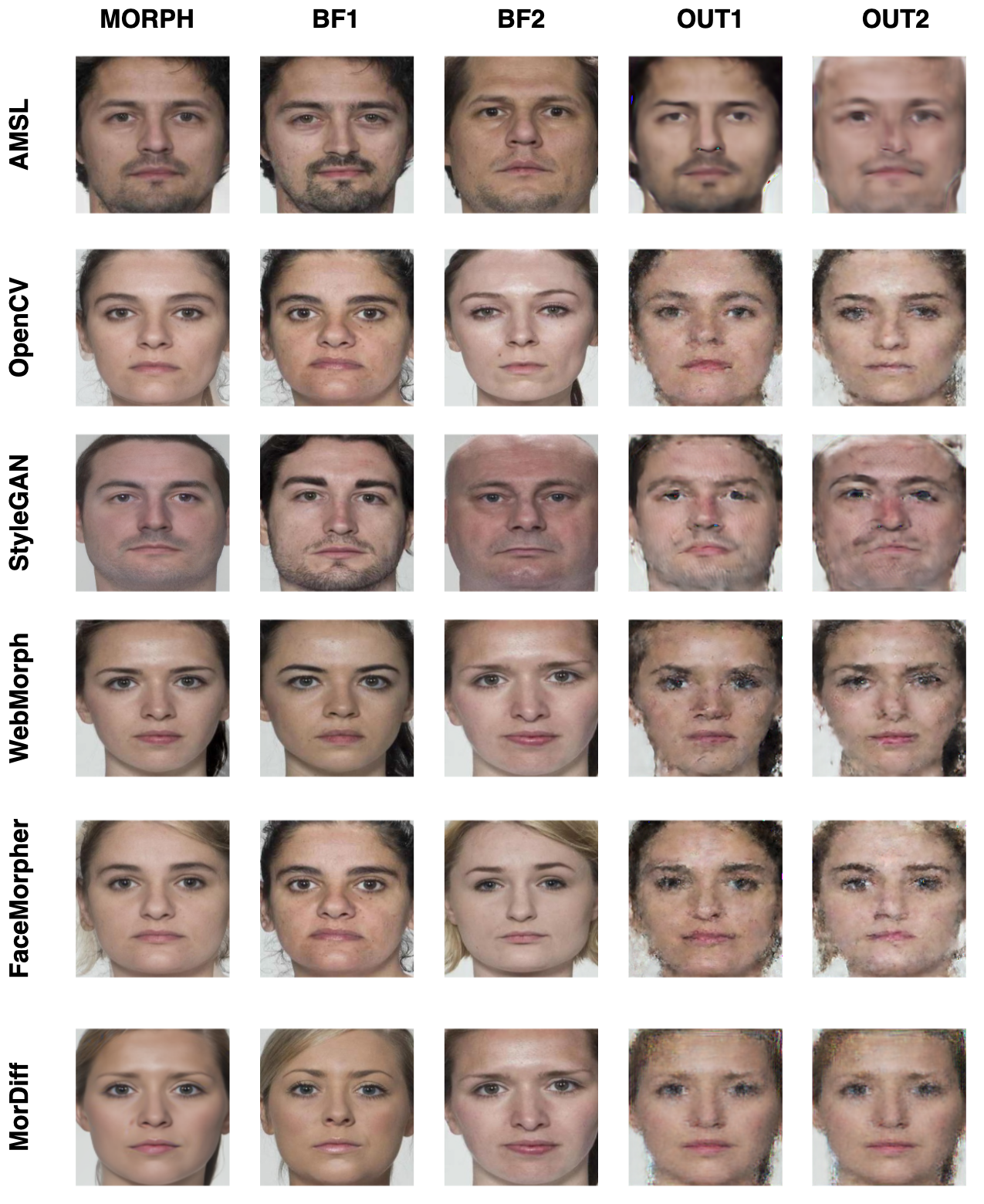}
    \caption{Reference-Free Demorphing Visualization: We illustrate the results of our reference-free demorphing approach. The model takes the morph image (MORPH) as input and produces two outputs (OUT1 and OUT2). The ground-truth images used to generate the morph are denoted as BF1 and BF2.}
    \label{fig:recon}
\end{figure}

\begin{figure}
    \centering
    \includegraphics[width=\linewidth]{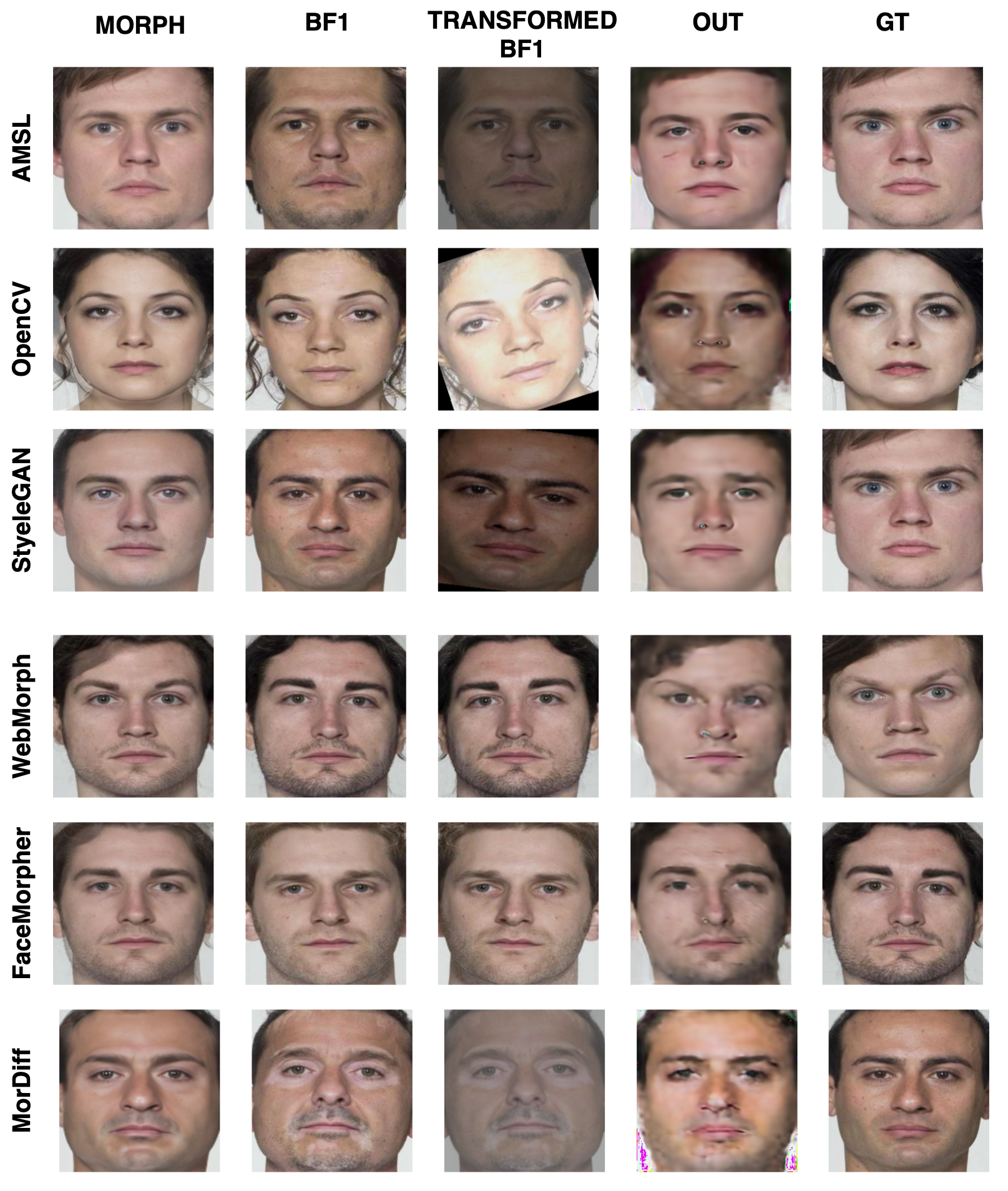}
    \caption{Differential Demorphing Visualization: We illustrate the results of our differential demorphing approach. The model receives the morph image (MORPH) and the transformed ground-truth image (TRANSFORMED BF1) and produces an output (OUT). The ground truth (GT) is shown, which, along with the ground-truth image (BF1), was used to generate the morph image (MORPH). Note that our method does not assume that the demorphing process uses the exact same ground-truth image (BF1) that was used to create the morph.   }
    \label{fig:diff-recon}
\end{figure}

\begin{table}[h!]
\centering
\caption{We compute image quality between generated face images and ground truth face images.}
\resizebox{0.9\columnwidth}{!}{
\begin{tabular}{|l|c|c|c|c|}
\hline
\textbf{Method} & \textbf{FID}$\downarrow$ & \textbf{SSIM}$\uparrow$ & \textbf{LPIPS}$\downarrow$ & \textbf{PSNR}$\uparrow$ \\
\hline
AMSL          & 0.3816 & 0.7765 & 0.2356 & 18.1377 \\
OpenCV        & 0.1976 & 0.5742 & 0.2318 & 17.7306 \\
FaceMorpher   & 0.5853 & 0.5554 & 0.2459 & 17.7578 \\
WebMorph      & 0.2258 & 0.5696 & 0.2277 & 17.7022 \\
StyleGAN      & 0.1506 & 0.5607 & 0.2520 & 18.3152 \\
MorDiFF            & 1.0297 & 0.6631 & 0.2552 & 18.5223 \\
\hline
\textbf{Average}   & \textbf{0.4284} & \textbf{0.6166} & \textbf{0.2414} & \textbf{18.0276} \\
\hline
\end{tabular}
}
\label{tab:IQA}
\end{table}
\subsection{Findings}

We evaluate our method both visually and analytically. It is imperative for any demorphing method to produce visually distinctive faces compared to the input morph as well as between themselves. Note that a trivial solution  for demorphing would be an identity function which simply outputs the morph. By construction, the outputs will have high similarity scores with the bonafide faces used to create the morph in the database, rendering the demorphing process unnecessary. Therefore, visual evaluation plays a crucial role in demonstrating the effectiveness of our method. 

\subsubsection{Reference-Free Demorphing} 
\begin{figure}
    \centering
    \includegraphics[width=0.8\linewidth]{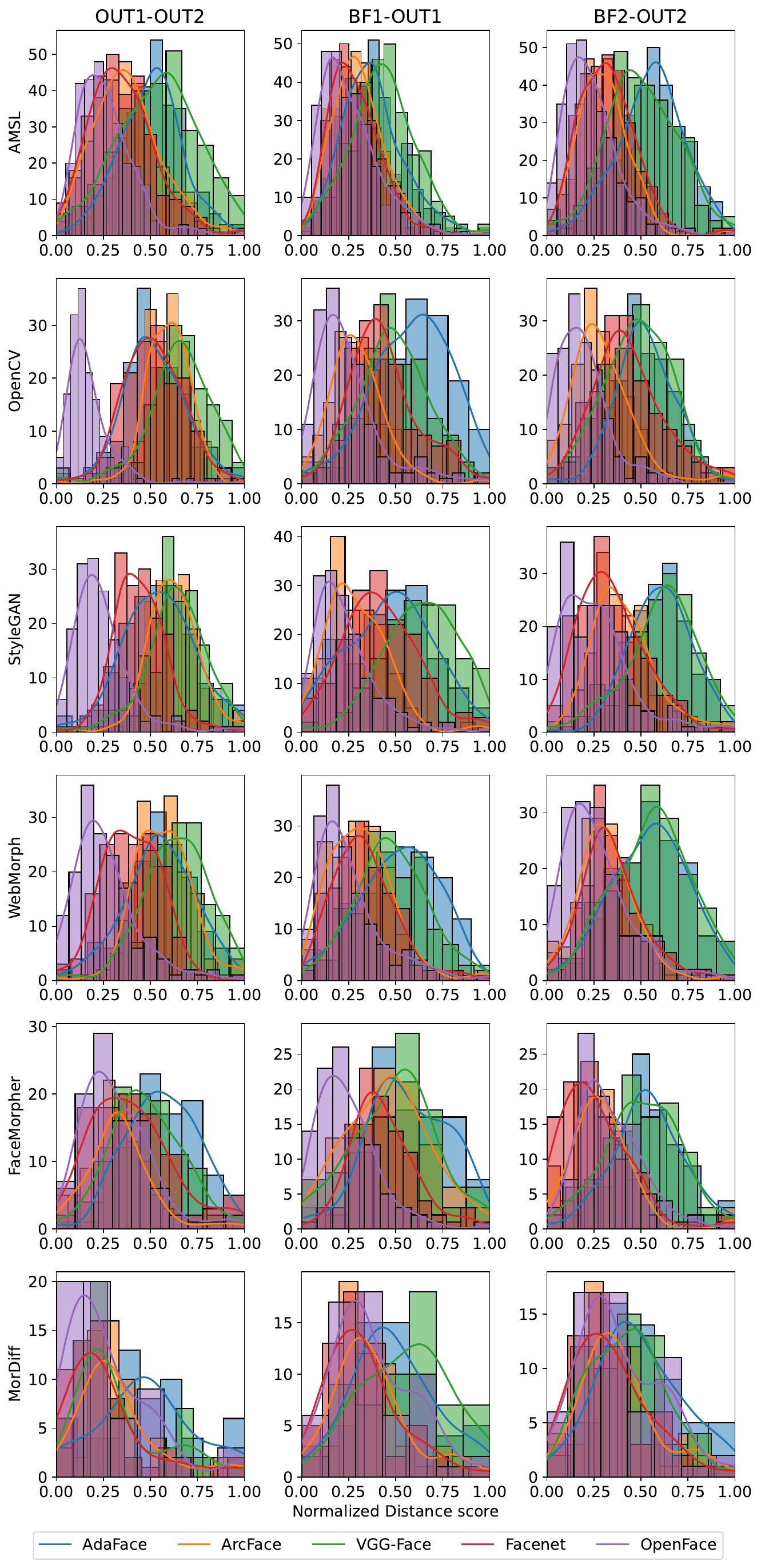}
    \caption{We plot the normalized distance scores for the pairs OUTPUT1-OUTPUT2, INPUT1-OUTPUT1, and INPUT2-OUTPUT2. Our method produces distinct face images, as illustrated in the left column.}
    \label{fig:similarity-dis}
\end{figure}

We visualize the outputs of our reference-free demorphing method across six morphing techniques in Figure \ref{fig:recon}. These techniques include both landmark-based and deep-learning-based morphs. The first three columns visualize the morph and the two constituent face images used to create it, while the subsequent two columns show the outputs of our method, presented in no specific order. We observe that the faces produced by our method are visually distinct from each other and also appear notably different from the original morph. Moreover, they are visually closer to their ground-truth counterparts. To quantify this, we calculate the normalized biometric {\em distance} between the generated faces and between the generated faces and the ground truth. We plot the distance scores across three different pairings of images: OUT1-OUT2, BF1-OUT1, and BF2-OUT2, which represent the dissimilarity between the generated outputs, and between each generated output and its corresponding ground-truth. We plot the distance distribution for all datasets across all FR models in Figure \ref{fig:similarity-dis}. We observe an overall OUT1-OUT2 biometric distance of 0.4380, indicating a quantifiable difference in the output faces. Note that we use cosine distance rather than similarity to quantify results. This choice helps us intuitively explain \textit{morph-replication} because a greater distance indicates that the generated faces are more distinguishable from each other. For BF1-OUT1, we observe a distance of 0.3992, and for BF2-OUT2, a distance of 0.4025. Note that the distance scores between outputs and their respective ground-truth face images fall well below the threshold used in several biometric systems \cite{ref82,ref83}. Our method successfully learns to distinguish features from the base image, such as hair (Rows 1 and 3), skin color, and subtle details like eyebrows and facial hair. Finally, we compute the True Match Rate (TMR) of the output subjects at various False Match Rate (FMR) thresholds and present the results in Table \ref{tab:performance_comparison}. Our method achieves an average TMR of 93.86\% on the AMSL dataset. The corresponding TMR values on the OpenCV, StyleGAN, WebMorph, FaceMorph, and MorDiff datasets are 93.99\%, 58.92\%, 89.87\%, 94.39\%, and 93.75\%, respectively.
We also evaluate the quality of images generated by our method using various image quality assessment (IQA) metrics, as shown in Table \ref{tab:IQA}. Our method achieves an average FID \cite{ref84} of 0.4284 across all datasets, with corresponding averages of 0.61 for SSIM \cite{ssim}, 0.24 for LPIPS \cite{lpips}, and 18.02 for PSNR, demonstrating the visual realism of the outputs produced by our method. We also observe that ArcFace performs consistently across all datasets - despite the presence of visual artifacts generated during demorphing - compared to other face models, which is in line with \cite{ref51}.

\noindent\textbf{Comparison with SOTA: } We evaluate our method against existing reference-free demorphing approaches based on restoration accuracy and TMR at 10\% FMR. Restoration accuracy measures the proportion of outputs that correctly match their corresponding ground-truth images. Our method achieves an average TMR of 93.86\% on the AMSL dataset using the ArcFace model, surpassing the performance of \cite{ref51} by 23.32\%. Reference-free demorphing methods, particularly for scenario 3, remain limited. Nonetheless, we compare our method's restoration accuracy with \cite{ref18,ref66} in both scenarios 1 and 3. In scenario 3, our method shows a significant performance advantage and achieves comparable results in scenario 1, despite being trained in the more challenging scenario 3. Results are summarized in Table \ref{tab:compare}.

\noindent\textbf{Effect of $\mathcal{L}_{cr}$ and $\mathcal{E}(\cdot)$:} We conduct an ablation study on our method to assess i) the impact of cross-road loss and ii) the influence of dual conditioning on output generation. Excluding pixel-wise guidance (the second part of the loss in Equation \ref{loss}) results in an average performance decline of 20.87\% across the six datasets tested using the ArcFace model. This decrease is mainly due to the method's reduced ability to accurately align the two unordered generated outputs with the ordered ground-truth images. Moreover, when the generator is trained without $\mathcal{E}$, the performance drops by 8.57\% due to lack of context during the generation process (see Section \ref{proposed}). These results are presented in Table \ref{tab:my_label}.

\subsubsection{Differential demorphing} We visualize the results of differential demorphing in Figure \ref{fig:diff-recon}. The first two columns show MORPH and BF1, the third column displays the randomly transformed BF1 used as the reference image in the demorphing process, and the remaining two columns present the output (OUT) from our method and the corresponding ground truth (GT).
Our method produces faces that are visibly distinct from both the MORPH and the reference image, achieving a TMR of 94.88\% on the AMSL dataset, with corresponding values of 88.12\%, 54.04\%, 89.38\%, 98.0\%, and 90.32\% on the OpenCV, StyleGAN, WebMorph, FaceMorph, and MorDiff datasets, respectively. These results are presented in Table \ref{tab:performance_comparison_diff}.

\begin{table}[htbp]
    \centering
    \caption{We calculate the average biometric distance for three cases — OUT1-OUT2, BF1-OUT1, and BF2-OUT2 — across all five FR models utilized. }
    \resizebox{\columnwidth}{!}{
    \begin{tabular}{|c|c|c|c|}
        \hline
        \textbf{Method} & \textbf{OUT1-OUT2} $\uparrow$ & \textbf{BF1-OUT1} $\downarrow$ & \textbf{BF2-OUT2}$\downarrow$ \\ \hline
        AMSL  & 0.4053 & 0.3324 & 0.3847 \\ \hline
        OpenCV & 0.4910 & 0.4115 & 0.3878 \\ \hline
        FaceMorpher & 0.4961 & 0.4084 & 0.4442 \\ \hline
        WebMorph & 0.4825 & 0.3802 & 0.4078 \\ \hline
        StyleGAN & 0.4193 & 0.4377 & 0.3864 \\ \hline
        MorDiff & 0.3338 & 0.4247 & 0.4043 \\ \hline
        \textbf{Average} & \textbf{0.4380} & \textbf{0.3992} & \textbf{0.4025} \\ \hline
    \end{tabular}}
    
    \label{tab:means_table}
\end{table}




\begin{table}[h!]
\centering
\caption{We calculate the $d'$ values for Genuine versus Imposter scores in differential demorphing. Higher $d'$ values suggest greater separability between the distributions.}
\resizebox{\columnwidth}{!}{%
\begin{tabular}{|c|c|c|c|c|c|}
\hline
 & AdaFace & ArcFace & VGG-Face & FaceNet & OpenFace \\ \hline
AMSL        & 0.22 & \textbf{2.66} & 0.07 & 0.11 & 0.29 \\ \hline
OpenCV      & \textbf{2.73} & 3.60 & 0.82 & 1.34 & 0.09 \\ \hline
StyleGAN    & 1.39 & 0.80 & \textbf{1.17} & 1.02 & 0.99 \\ \hline
WebMorph    & 0.93 & \textbf{2.07} & 0.10 & 0.27 & 0.15 \\ \hline
FaceMorpher & 1.49 & \textbf{2.04} & 0.35 & 0.50 & 0.33 \\ \hline
MorDiff     & \textbf{2.02} & 1.96 & 0.62 & 0.45 & 0.17 \\ \hline
\end{tabular}
}
\end{table}

\begin{table}[h!]
    \centering
      \caption{Effect of the dual-condition and $\mathcal{L}_{cr}$ across the five FR models used (AdaFace/ArcFace/VGG-Face/FaceNet/OpenFace). The drop in TMR performance shows the importance of the pixel-wise guidance during training and the effectiveness of the condition by $\mathcal{E}$ during output generation. }
    \resizebox{!}{0.12\columnwidth}{
    \begin{tabular}{|c|c|c|}
    \hline
         Dataset & w/o $\mathcal{L}_{cr}$ & w/o $\mathcal{E}(\cdot)$ \\
         \hline
         AMSL & 32.71/74.49/37.50/11.99/15.75 & 24.14/82.88/5.48/9.59/9.34 \\
         OpenCV & 42.81/92.79/25.49/27.12/15.03 & 57.86/86.65/29.56/13.52/15.03 \\
         StyleGAN & 9.03/56.13/7.42/7.74/12.58 & 6.25/52.5/6.25/9.38/3.75 \\
         WebMorph & 39.69/84.38/28.75/29.69/17.19 & 36.25/79.69/25.31/16.56/10.0 \\
         FaceMorpher & 43.03/91.30/26.77/19.19/22.73 & 50.00/90.04/32.83/25.25/9.60 \\
         MorDiff & 37.29/0.0/1.69/3.39/15.25 & 48.86/81.25/40.18/24.11/21.43 \\
         \hline
    \end{tabular}}
  
    \label{tab:my_label}
\end{table}

\section{Summary}
\label{summary}

In this paper, we propose dc-GAN, a novel method to recover constituent images from a facial morph. Our proposed method is based on the GAN architecture; we condition our GAN on morphs as well as their latent representations to provide robust guidance while recovering constituent images. Previous approaches suffered from \textit{morph-replication}, i.e., they produced similar looking faces as the morph itself. Our method significantly overcomes this problem. We evaluate our method across 6 different morphing techniques using five biometric face models. We assess our method using the AMSL, FRLL-Morphs, and MorDiff datasets, achieving visually compelling results and addressing the issue of morph-replication. Moreover, experiments suggest that ArcFace performs consistently for demorphing tasks compared to other face recognition models. Future work will aim to develop techniques for demorphing faces with various styles, beyond just passport-style face images.

\section{Acknowledgment}
This work was supported in part by the NSF Center for Identification Technology Research (CITeR) and the DHS Center for Criminal Investigations and Network Analysis (CINA). 
\\
\\
\noindent {\em This research did not involve the collection of data from human subjects. All experiments were carried out using existing face datasets available for research purposes, and no attempts were made to identify individuals within these datasets. Additionally, no personally identifiable information (PII) was used during the demorphing process described in this paper.}

{\small
\balance
\bibliographystyle{ieee_fullname}
\bibliography{egbib}

\begin{thebibliography}{10}\itemsep=-1pt

\bibitem{ref23}
{GNU} image manipulation program (gimp), 2016.

\bibitem{ref41}
P. Aghdaie, B. Chaudhary, S. Soleymani, J. Dawson, and N.~M. Nasrabadi.
\newblock Morph detection enhanced by structured group sparsity.
\newblock In {\em IEEE/CVF Winter Conference on Applications of Computer Vision Workshops (WACVW)}. IEEE Computer Society, 2022.

\bibitem{ref80}
Brandon Amos, Bartosz Ludwiczuk, and Mahadev Satyanarayanan.
\newblock Openface: A general-purpose face recognition library with mobile applications.
\newblock Technical report, CMU-CS-16-118, CMU School of Computer Science, 2016.

\bibitem{ref14}
André Anjos and Sébastien Marcel.
\newblock Counter-measures to photo attacks in face recognition: A public database and a baseline.
\newblock In {\em International Joint Conference on Biometrics (IJCB)}, pages 1--7, 2011.

\bibitem{ref51}
Sudipta Banerjee, Prateek Jaiswal, and Arun Ross.
\newblock Facial de-morphing: Extracting component faces from a single morph.
\newblock In {\em IEEE International Joint Conference on Biometrics (IJCB)}, 2022.

\bibitem{9484355}
Sudipta Banerjee and Arun Ross.
\newblock Conditional identity disentanglement for differential face morph detection.
\newblock In {\em 2021 IEEE International Joint Conference on Biometrics (IJCB)}, pages 1--8, 2021.

\bibitem{ref9}
Naser Damer, Meiling Fang, Patrick Siebke, Jan~Niklas Kolf, Marco Huber, and Fadi Boutros.
\newblock Mordiff: Recognition vulnerability and attack detectability of face morphing attacks created by diffusion autoencoders.
\newblock In {\em 11th International Workshop on Biometrics and Forensics (IWBF)}, pages 1--6, 2023.

\bibitem{ref20}
Naser Damer, C{\'e}sar Augusto~Fontanillo L{\'o}pez, Meiling Fang, No{\'e}mie Spiller, Minh~Vu Pham, and Fadi Boutros.
\newblock Privacy-friendly synthetic data for the development of face morphing attack detectors.
\newblock In {\em Proceedings of the IEEE/CVF Conference on Computer Vision and Pattern Recognition Workshop}, 2022.

\bibitem{ref70}
Lisa DeBruine.
\newblock debruine/webmorph: Beta release 2, 2018.

\bibitem{ref65}
Lisa DeBruine and Benedict Jones.
\newblock {Face Research Lab London Set}.
\newblock 5 2017.

\bibitem{ref77}
Jiankang Deng, Jia Guo, Niannan Xue, and Stefanos Zafeiriou.
\newblock Arcface: Additive angular margin loss for deep face recognition.
\newblock In {\em Proceedings of the IEEE/CVF Conference on Computer Vision and Pattern Recognition}, pages 4690--4699, 2019.

\bibitem{ref13}
Zeev Farbman, Raanan Fattal, Dani Lischinski, and Richard Szeliski.
\newblock Edge-preserving decompositions for multi-scale tone and detail manipulation.
\newblock {\em ACM Trans. Graph.}, 2008.

\bibitem{ref16}
Matteo Ferrara, Annalisa Franco, and Davide Maltoni.
\newblock Face demorphing.
\newblock {\em IEEE Transactions on Information Forensics and Security}, 2018.

\bibitem{ref21}
Matteo Ferrara, Annalisa Franco, and Davide Maltoni.
\newblock Decoupling texture blending and shape warping in face morphing.
\newblock In {\em 2019 International Conference of the Biometrics Special Interest Group (BIOSIG)}, pages 1--5, 2019.

\bibitem{ref32}
Ian~J. Goodfellow, Jean Pouget-Abadie, Mehdi Mirza, Bing Xu, David Warde-Farley, Sherjil Ozair, Aaron Courville, and Yoshua Bengio.
\newblock Generative adversarial nets.
\newblock In {\em Proceedings of the 27th International Conference on Neural Information Processing Systems}, 2014.

\bibitem{ref81}
Sylvain Gugger, Lysandre Debut, Thomas Wolf, Philipp Schmid, Zachary Mueller, Sourab Mangrulkar, Marc Sun, and Benjamin Bossan.
\newblock Accelerate: Training and inference at scale made simple, efficient and adaptable.
\newblock \url{https://github.com/huggingface/accelerate}, 2022.

\bibitem{ref84}
Martin Heusel, Hubert Ramsauer, Thomas Unterthiner, Bernhard Nessler, and Sepp Hochreiter.
\newblock {GAN}s trained by a two time-scale update rule converge to a local nash equilibrium.
\newblock In {\em {Proceedings of the 31st International Conference on Neural Information Processing Systems}}, page 6629–6640, 2017.

\bibitem{ref72}
Phillip Isola, Jun-Yan Zhu, Tinghui Zhou, and Alexei~A Efros.
\newblock Image-to-image translation with conditional adversarial networks.
\newblock {\em Proceedings of Computer Vision and Pattern Recognition}, 2017.

\bibitem{ref69}
Tero Karras, Samuli Laine, Miika Aittala, Janne Hellsten, Jaakko Lehtinen, and Timo Aila.
\newblock Analyzing and improving the image quality of {StyleGAN}.
\newblock In {\em Proceedings of IEEE/CVF Conference on Computer Vision and Pattern Recognition}, 2020.

\bibitem{ref22}
Minchul Kim, Anil~K Jain, and Xiaoming Liu.
\newblock Adaface: Quality adaptive margin for face recognition.
\newblock In {\em Proceedings of the IEEE/CVF Conference on Computer Vision and Pattern Recognition}, 2022.

\bibitem{ref60}
Diederik~P. Kingma and Jimmy Ba.
\newblock Adam: {A} method for stochastic optimization.
\newblock In Yoshua Bengio and Yann LeCun, editors, {\em 3rd International Conference on Learning Representations}, 2015.

\bibitem{ref67}
Satya Mallick.
\newblock Face morph using opencv—c++/python.
\newblock {\em LearnOpenCV}, 1(1), 2016.

\bibitem{ref11}
Matthias Monroy.
\newblock Laws against morphing, 2020.
\newblock \url{https://digit.site36.net/2020/01/10/laws-against-morphing/}.

\bibitem{ref64}
Tom Neubert, Andrey Makrushin, Mario Hildebrandt, Christian Kraetzer, and Jana Dittmann.
\newblock Extended stirtrace benchmarking of biometric and forensic qualities of morphed face images.
\newblock {\em IET Biometrics}, 2018.

\bibitem{ref10}
Mei Ngan, Patrick Grother, Kayee Hanaoka, and Jason Kuo.
\newblock {Face Recognition Vendor Test (FRVT) Part 4: MORPH - Performance of Automated Face Morph Detection}, March 2020.

\bibitem{ref78}
Omkar~M. Parkhi, Andrea Vedaldi, and Andrew Zisserman.
\newblock Deep face recognition.
\newblock In {\em Proceedings of the British Machine Vision Conference}, 2015.

\bibitem{ref73}
Deepak Pathak, Philipp Kr\"ahenb\"uhl, Jeff Donahue, Trevor Darrell, and Alexei Efros.
\newblock Context encoders: Feature learning by inpainting.
\newblock In {\em Proceedings of the Conference on Computer Vision and Pattern Recognition}, 2016.

\bibitem{ref17}
Fei Peng, Le-Bing Zhang, and Min Long.
\newblock {FD-GAN: Face de-morphing generative adversarial network for restoring accomplice’s facial image}.
\newblock {\em IEEE Access}, 2019.

\bibitem{ref68}
A Quek.
\newblock Face morpher.
\newblock \url{ https://github.com/alyssaq/face_morpher}.

\bibitem{ref74}
Alec Radford, Jong~Wook Kim, Chris Hallacy, Aditya Ramesh, Gabriel Goh, Sandhini Agarwal, Girish Sastry, Amanda Askell, Pamela Mishkin, Jack Clark, Gretchen Krueger, and Ilya Sutskever.
\newblock Learning transferable visual models from natural language supervision.
\newblock In {\em International Conference on Machine Learning}, 2021.

\bibitem{ref50}
R. Raghavendra, Kiran~B. Raja, and Christoph Busch.
\newblock Detecting morphed face images.
\newblock In {\em IEEE 8th International Conference on Biometrics Theory, Applications and Systems (BTAS)}, 2016.

\bibitem{ref75}
Robin Rombach, Andreas Blattmann, Dominik Lorenz, Patrick Esser, and Bj{\"o}rn Ommer.
\newblock High-resolution image synthesis with latent diffusion models.
\newblock In {\em Proceedings of the IEEE/CVF Conference on Computer Vision and Pattern Recognition}, 2022.

\bibitem{ref19}
Olaf Ronneberger, Philipp Fischer, and Thomas Brox.
\newblock U-net: Convolutional networks for biomedical image segmentation.
\newblock In {\em Medical Image Computing and Computer-Assisted Intervention (MICCAI)}, 2015.

\bibitem{ref53}
Ulrich Scherhag, Dhanesh Budhrani, Marta Gomez-Barrero, and Christoph Busch.
\newblock Detecting morphed face images using facial landmarks.
\newblock In Alamin Mansouri, Abderrahim El~Moataz, Fathallah Nouboud, and Driss Mammass, editors, {\em Image and Signal Processing}, Cham, 2018. Springer International Publishing.

\bibitem{ref79}
Florian Schroff, Dmitry Kalenichenko, and James Philbin.
\newblock Facenet: A unified embedding for face recognition and clustering.
\newblock In {\em Proceedings of the IEEE Conference on Computer Vision and Pattern Recognition (CVPR)}, June 2015.

\bibitem{ref83}
Sefik Serengil and Alper Ozpinar.
\newblock A benchmark of facial recognition pipelines and co-usability performances of modules.
\newblock {\em Journal of Information Technologies}, 17(2):95--107, 2024.

\bibitem{ref82}
Sefik~Ilkin Serengil and Alper Ozpinar.
\newblock Lightface: A hybrid deep face recognition framework.
\newblock In {\em Innovations in Intelligent Systems and Applications Conference (ASYU)}, pages 23--27, 2020.

\bibitem{ref18}
Nitish Shukla.
\newblock Sdemorph: Towards better facial de-morphing from single morph.
\newblock In {\em IEEE International Joint Conference on Biometrics (IJCB)}, 2023.

\bibitem{ref66}
Nitish Shukla and Arun Ross.
\newblock Facial demorphing via identity preserving image decomposition.
\newblock {\em IEEE International Joint Conference on Biometrics (IJCB)}, 2024.

\bibitem{ref76}
Dmitry Ulyanov, Andrea Vedaldi, and Victor~S. Lempitsky.
\newblock Instance normalization: The missing ingredient for fast stylization.
\newblock {\em ArXiv}, abs/1607.08022, 2016.

\bibitem{ssim}
Zhou Wang, A.C. Bovik, H.R. Sheikh, and E.P. Simoncelli.
\newblock Image quality assessment: from error visibility to structural similarity.
\newblock {\em IEEE Transactions on Image Processing}, 13(4):600--612, 2004.

\bibitem{ref71}
K. Zhang, Z. Zhang, Z. Li, and Y. Qiao.
\newblock Joint face detection and alignment using multitask cascaded convolutional networks.
\newblock {\em IEEE Signal Processing Letters}, 23(10):1499--1503, October 2016.

\bibitem{lpips}
Richard Zhang, Phillip Isola, Alexei~A Efros, Eli Shechtman, and Oliver Wang.
\newblock The unreasonable effectiveness of deep features as a perceptual metric.
\newblock In {\em Proceedings of the IEEE Conference on Computer Vision and Pattern Recognition (CVPR)}, 2018.

\end{thebibliography}
}

\end{document}